\documentclass[sigconf,prologue,table]{acmart}
\usepackage{multirow}
\usepackage{algorithm}
\usepackage{algpseudocode}

\AtBeginDocument{%
  \providecommand\BibTeX{{%
    \normalfont B\kern-0.5em{\scshape i\kern-0.25em b}\kern-0.8em\TeX}}}

\acmSubmissionID{2305}

\begin{document}

\title{
Double Mixture: Towards Continual Event Detection from Speech
}

\newcommand{\monash}{\spadesuit}
\newcommand{\seu}{\heartsuit}
\author{Jingqi Kang$^{*,\seu}$\ \ \ \ \ Tongtong Wu$^{*,\monash}$\ \ \ \ \  Jinming Zhao$^{\monash}$\ \ \ \ \  Guitao Wang$^\seu$\ \ \ \ \ Yinwei Wei$^\monash$\ \ \ \ \ Hao Yang$^\monash$ \\
{Guilin Qi}$^{\seu}$\ \ \ \ \ {Yuan-Fang Li}$^{\monash}$\ \ \ \ \ {Gholamreza Haffari}$^{\monash}$ \\ [4pt]
$^{\seu}$Southeast University, China; $^{\monash}$Monash University, Australia \\[4pt]
$^{\seu}$\texttt{\{kjq, 220222117, gqi\}@seu.edu.cn}, \\[4pt]
$^{\monash}$\texttt{\{first-name.last-name\}@monash.edu}\\[4pt]
}


\renewcommand{\shortauthors}{Kang and Wu, et al.}


\begin{abstract}
  Speech event detection is crucial for multimedia retrieval, involving the tagging of both semantic and acoustic events. Traditional ASR systems often overlook the interplay between these events, focusing solely on content, even though the interpretation of dialogue can vary with environmental context. This paper tackles two primary challenges in speech event detection: the continual integration of new events without forgetting previous ones, and the disentanglement of semantic from acoustic events. We introduce a new task, continual event detection from speech, for which we also provide two benchmark datasets. To address the challenges of catastrophic forgetting and effective disentanglement, we propose a novel method, 'Double Mixture.' This method merges speech expertise with robust memory mechanisms to enhance adaptability and prevent forgetting. Our comprehensive experiments show that this task presents significant challenges that are not effectively addressed by current state-of-the-art methods in either computer vision or natural language processing. Our approach achieves the lowest rates of forgetting and the highest levels of generalization, proving robust across various continual learning sequences. Our code and data are available at \url{https://github.com/jodie-kang/DoubleMixture}.
\end{abstract}

%







\maketitle

\section{Introduction}
Speech event detection is vital for multimedia retrieval~\cite{wang2023understanding}, as it forms the basis for precise indexing and accessing of extensive audio content~\cite{gong2023whisper, kang2024event}. This task involves recognizing not just the semantic content of dialogues, essentially the words spoken, but also the acoustic events, which indicate the location and background conditions of the speech. Current information extraction systems~\cite{WuWZLQLH22,kang2024event} based on automatic speech recognition (ASR) systems~\cite{gong2023whisper} often fail to address the interaction between semantic and acoustic signals. This oversight is particularly problematic in complex multimedia environments where background sounds and speech frequently overlap~\cite{deldjoo2020recommender,yi2019acm}.

To address the challenges in extracting events from real-world speech data, the models need to meet two key criteria: continual learning~\cite{ChangLL21} and event disentanglement~\cite{LuW0M23}. Continual learning enables models to adapt to an evolving data landscape, improving their ability to recognize new event types while retaining previously learned information. Disentangling semantic events from their acoustic environments allows models to handle rare or previously unseen event combinations effectively.

Current research typically approaches speech-based information extraction as an extension of text-based methods~\cite{kang2024event, WuCLLQH22}, focusing predominantly on end-to-end extraction from a content perspective. This approach often neglects the unique challenges of speech data. In response, we propose a new task: Continual Event Detection from Speech (CEDS). Due to a scarcity of specific datasets, we suggest utilizing existing text-based and acoustic event detection datasets to establish a new benchmark for continual learning, employing methods such as speech overlaying and speech splicing.

\begin{figure}
\includegraphics[width=0.475\textwidth]{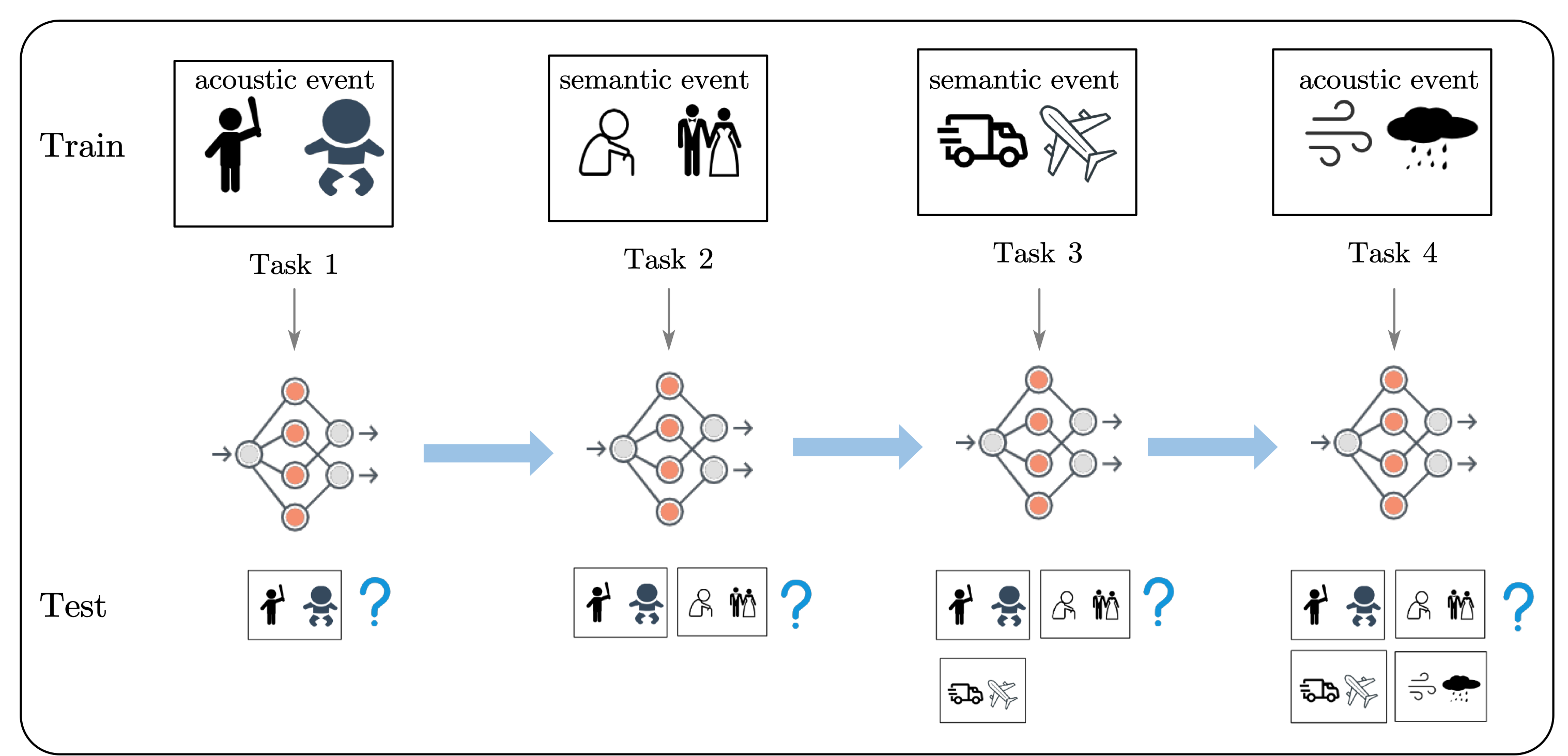}
  \caption{In continual learning, learners incrementally acquire new event types and must evaluate all previously learned types during testing. This process is particularly challenging in speech-based scenarios due to the complex interplay of semantic content (semantic event) and background sounds (acoustic event).\label{fig:problem}}
\end{figure}

To tackle the challenges of continual learning and event disentanglement in speech, we introduce a novel strategy called Double Mixture. This approach combines a mixture of experts~\cite{riquelme2021scaling, zhou2022mixture, chen2023mod} with automatically assigning a dedicated expert to each task for accruing new knowledge, and a mixture of memory, which is a simple yet effective method for replaying speech experiences. The proposed method aims to prevent catastrophic forgetting and improve the model's ability to process complex audio inputs.
In summary, our contributions are as follows:

\begin{figure*}[!htbp]
  \includegraphics[width= \textwidth]{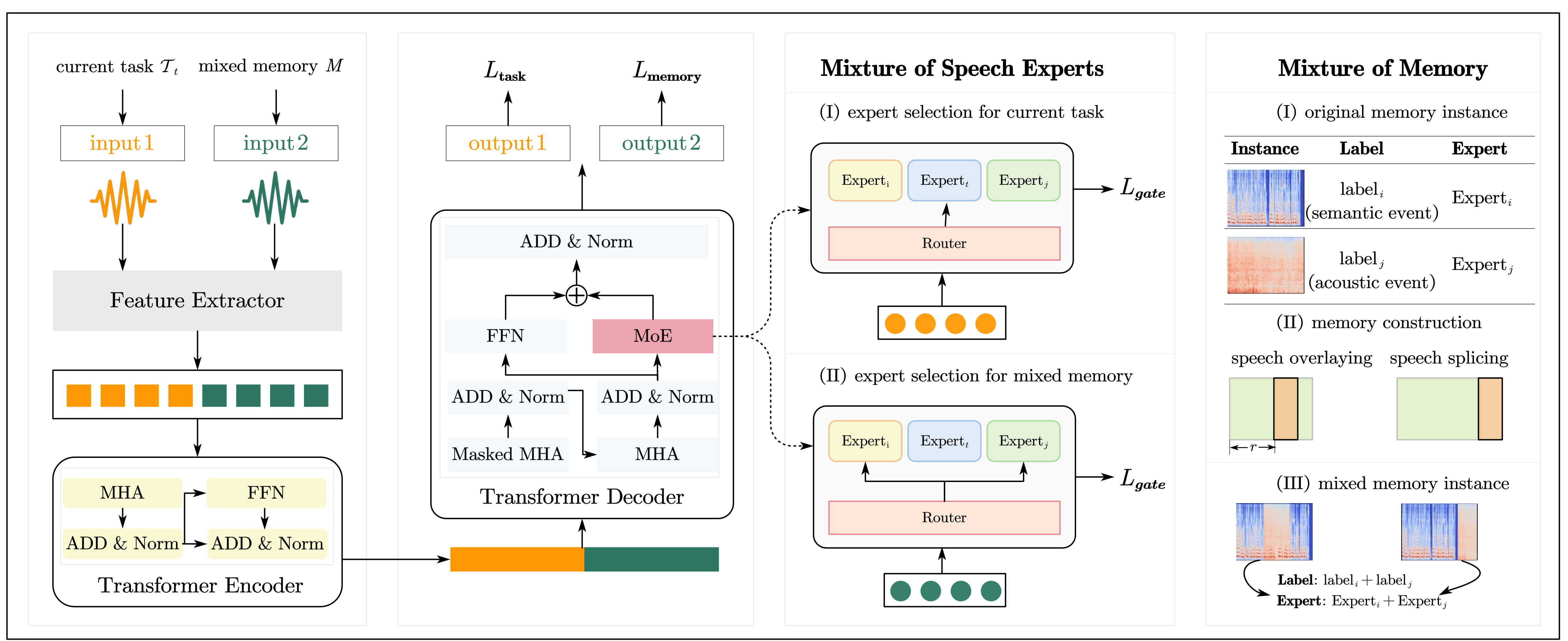}
  \caption{Framework of the proposed Double Mixture method.}
  \label{fig:method}
\end{figure*}
\begin{enumerate}
    \item We are the first to propose the task of continual event detection from speech, providing new benchmarks and highlighting the significant challenges of catastrophic forgetting and semantic-acoustic event disentanglement.
    \item We introduce Double Mixture, a method that merges a mixture of experts with a mixture of memory, to enhance performance in complex audio scenarios.
    \item Our extensive experiments confirm the complexity of this task and demonstrate the state-of-the-art performance of our proposed method in preventing catastrophic forgetting and managing complex real-world audios with varying event combinations.
\end{enumerate}

\section{Task Definition}
The continual speech event detection task aims to sequentially learn and recognize new tasks from a speech stream, and the process can be formally defined as follows:

Consider a task sequence $\mathcal{T}=\left \{ \mathcal{T}_1, \dots, \mathcal{T}_t, \dots, \mathcal{T}_T \right \}$, where $T$ denotes the sequence length. Each task $\mathcal{T}_t= \{ \mathcal{D}_t^{train}, \mathcal{D}_t^{val}, \mathcal{D}_t^{test} \} $ in the sequence consists of a supervised learning task including a training set $D_t^{train}$, a validation set $\mathcal{D}_t^{val}$ and a test set $\mathcal{D}_t^{test}$.
Each dataset $\mathcal{D}$ contains $n$ sample pairs $\mathcal{D}={(x_i, y_i)}_{i=1}^n$, where $x_i$ is a sequence of digital audio signals, and $y_i \in \mathcal{Y}$ denotes the event type. The event type can be either a semantic event or a sound event.
Assuming that $\mathcal{Y}^{train}_t$ represents the set of event types occurring in $\mathcal{D}^{train}_t$, the incremental speech event extraction task requires that the sets of event types do not intersect with each other during training, i.e., $\varnothing = \bigcap_{t=1}^T\mathcal{Y}^{train }_t$. However, the set of event types at the test phase is $\mathcal{Y}^{test}_T = \bigcup_{t=1}^T \mathcal{Y}^{test}_t$, allowing the model to encounter all previously learned event types throughout the test phase of the task sequence.

The catastrophic forgetting problem is the degradation of the model's performance on the previous task after learning a new task.
The event disengagement problem refers to combining the knowledge learned in previous tasks and applying it to new tasks.



\section{Method}
Our method employs the encoder-decoder Transformer\cite{vaswani2017attention}, including a speech encoder and a text decoder.
Inspired by the Mixed of Experts model (MoE)~\cite{shazeer2017outrageously, riquelme2021scaling}, we introduce a mixture of speech expert networks in the decoder to mitigate catastrophic forgetting in continual learning.
Different from the traditional experience replay strategy, we save mixed samples of semantic and sound events in memory to strengthen collaboration between different speech experts and thereby improve the generalization ability of the model.

\subsection{Mixture of Speech Experts}
The key issue in continual learning is how to avoid the loss of past knowledge while the model continues to absorb new knowledge.
MoE models are shown effective in mitigating the issue in the way that they combine the strength of multiple expert systems to achieve an in-depth understanding of specific tasks and work together in the overall performance~\cite{fedus2022switch}.
We thus introduce the mixture of speech experts in our framework where each expert focuses on a single task in the task stream, while the overall model responds optimally to new data by dynamically adjusting the weights of each expert,
and maintaining past knowledge during the learning process. 
As will be demonstrated in later sections, this method can improve the generalization ability of the model, i.e., it can use the learned event knowledge to identify new event combinations.

We design each speech expert as a bottleneck adapter network, which consists of a down-projection layer $W_{down}$ for dimensionality reduction, a nonlinear activation layer $\sigma(\cdot)$, and an up-projection layer $W_{up}$ for dimensionality restoration:
\begin{equation}
    E(\mathbf{H}) = \mathbf{H} + \sigma (\mathbf{H} \times W_{down}) W_{up},
    \label{eq:adapter}
\end{equation}
where $\mathbf{H} \in \mathbb{R}^{(p \times d)}$ is the hidden layer representation from the $l$-th layer, where $d$ is the dimension of the hidden layer and $p$ is the sequence length.
$W_{\text{down}} \in \mathbb{R}^{d \times b}$ and $W_{\text{up}} \in \mathbb{R}^{b \times d}$ are the trainable parameters of the adapter layer, and $b$ is the bottleneck dimension.

The core function of the adapter is to enhance the model's ability to process specific features, which is achieved by adding a learnable transformation to the original input:
\begin{equation}
    A(\mathbf{x}) = \sigma(W_f \times \mathbf{x} + b_f) + \mathbf{x},
\end{equation}
where $A(\mathbf{x})$ represents the output of the adapter, $\mathbf{x}$ is the input feature, $W_f$ and $b_f$ are the weights and biases of the adapter, and $\sigma$ denotes the activation function.

These weights $\alpha_i$ are usually normalized to ensure that the weights of all the experts sum to one, thus keeping the size of the output constant:
\begin{equation}
    \alpha_i = \frac{e^{G_i(\mathbf{x})}}{\sum_{j=1}^{n} e^{G_j(\mathbf{x})}},
    \label{eq:moe_weight}
\end{equation}
where $G_i(\mathbf{x})$ is a learned function to compute the importance score of the $i$-th adapter concerning the input $x$, which is then normalized by a softmax function. 

The mixed speech expert system dynamically integrates the outputs of each adapter through a gating mechanism that allows the system to adaptively adjust the contribution of each adapter according to the current input features and contextual demands. This dynamic integration process can be represented as:
\begin{equation}
    O(\mathbf{x}) = \sum_{i=1}^{n} \alpha_i A_i(\mathbf{x}),
    \label{eq:moe_output}
\end{equation}
where $O(\mathbf{x})$ is the total output of the system, $A_i(x)$ is the output of the $i$-th adapter, and $\alpha_i$ is the dynamic weight determined by the gating function $G(\mathbf{x})$, which reflects the importance of the adapters under the current input.

\subsection{Mixture of Memory}
Traditional experience replay methods~\cite{rolnick2019experience,lopez2017gradient} retain a portion of old task samples and combine them with new task data to form a training set. We store mixed speech samples containing semantic events and acoustic events in memory, named mixed memory. We store these mixed samples in a memory buffer and use them with new data during model training.

Mathematically, it is assumed that the memory buffer $\mathcal{M}$ contains the selected set of samples ${(x_i, y_i)}$, where $x_i$ is the sample feature and $y_i$ is the corresponding label. In the training phase, the model learns while drawing samples from the memory buffer and the current task data. The data-centric loss function is:
\begin{equation}
    \mathcal{L}_{data} = \lambda \mathcal{L}_{task} + (1 - \lambda) \mathcal{L}_{memory},
\end{equation}
where $\mathcal{L}_{task}$ and $\mathcal{L}_{memory}$ are the cross-entropy loss, measuring the difference between the predicted probability distribution over the vocabulary and the actual distribution (one-hot encoded target) for instances from the current task data and the mixed memory respectively. $\lambda$ is a parameter used to balance the importance of old and new knowledge, and the value is 0.5 based on experience.

\subsection{Training Strategy}
During the training phase, we freeze the parameters of the speech encoder to preserve its ability to understand speech features and only fine-tune the parameters of the mixture of speech experts in the decoding layer.

We compute the loss of the gating network to optimize the gating mechanism assigned by experts, and the loss function is:
\begin{equation}
    \mathcal{L}_{gate} = -\sum_{j}^{|D|} t_j \cdot log\sum_{i=0}^{d}r_i,
\end{equation}
where $t_j$ is the gold task id of the $j$-th instance, $d$ denotes the total number of decoder blocks, and $r_i$ denotes the logits generated by the router.

Consequently, the aggregate training loss is formulated as a combination of the data-centric loss and the gating loss. The total loss is:
\begin{equation}
    \mathcal{L}_{total} = \mathcal{L}_{data} + \eta \mathcal{L}_{gate},
\end{equation}
where $\mathcal{L}_{data}$ focuses on the model's performance on the current training data, $\mathcal{L}_{gate}$ ensures that the gating mechanism effectively integrates the outputs of multiple speech experts, and $\eta$ is hyper-parameters used to balance out these two components of the loss.

\section{Experiment}

\subsection{Datasets}
We conduct experiments on three benchmark datasets, including the semantic event datasets Speech-ACE05 and Speech-MAVEN~\cite{kang2024event}, and the acoustic event dataset ESC-50~\cite{piczak2015esc}. 
We divide these datasets into a series of task sequences to satisfy continuous learning scenarios. Each task sequence corresponds to an independent speech event detection task to simulate the process of the model gradually being exposed to new tasks.
The detailed data statistics are shown in Table \ref{tab:dataset}.

\begin{table}[!htbp]
    \centering
    \resizebox{.48\textwidth}{!}{
        \begin{tabular}
        {l|c|c|c|c}
        \toprule
        \textbf{Dataset} & \textbf{\# Class} & \textbf{\# Task} & \textbf{\# Instance} & \textbf{\# Hours}\\
        \midrule
        Speech ACE05 & 20 & 7 & 13984 & 33.36  \\
        Speech MAVEN & 26 & 6 & 33666 & 81.28  \\
        ESC-50 & 50 & 5 & 2000 & 2.78  \\
        \bottomrule
        \end{tabular}
    }
    \caption{Dataset statistics. 
    The event types in the dataset are two-level, and we assign task categories according to the first-level labels. For example, “Life” includes “Be-born”, “Die”, “Injure” and “Marry”.
    }
    \label{tab:dataset}
\end{table}
\vspace{-.3cm}

To evaluate the generalization ability of the model in complex scenarios, we select the categories with a large number of samples from the above three speech data sets and then mix semantic events and acoustic events. Finally, we construct two new data sets, namely speech splicing (SS) and speech overlaying (SO).
Table \ref{tab:dataset-cg} shows their details.
\begin{table}[!htbp]
    \centering
    \resizebox{.48\textwidth}{!}{
        \begin{tabular}{c|c|c|c|c|c}
            \toprule
            \textbf{Task} & \textbf{Semantic} & \textbf{Sound} & \textbf{\# Instance} & \textbf{SS (Avg sec.)} & \textbf{SO (Avg sec)} \\
            \midrule
            1 & conflict & nature & 2169 & 9.18 & 6.16 \\ 
            2 & movement & nature & 1057 & 4.47 & 3.01 \\ 
            3 & scenario & animal & 6955 & 29.42 & 19.72 \\ 
            4 & talk & human & 4695 & 19.86 & 13.34 \\ 
            \bottomrule
        \end{tabular}
    }
    \caption{Statistics information of the two combined datasets, SS and SO are the abbreviations of Speech splicing and Speech overlay respectively, Avg sec. indicates the average duration seconds of each audio file.
    }
    \label{tab:dataset-cg}
\end{table}

\begin{table*}[!htbp]
    \centering
    \resizebox{.78\textwidth}{!}
    {
        \begin{tabular}{c|c|c|c|c|c|c}
            \toprule
            \multirow{2}{*}{\textbf{Method}} & \multicolumn{2}{c|}{\textbf{Speech ACE05}} & \multicolumn{2}{c|}{\textbf{Speech MAVEN}} & \multicolumn{2}{c}{\textbf{ESC-50}} \\
            \cline{2-7} 
            & Avg ACC$\uparrow$ & Avg Forgetting$\downarrow$ & Avg ACC$\uparrow$ & Avg Forgetting$\downarrow$ & Avg ACC$\uparrow$ & Avg Forgetting$\downarrow$ \\
            \midrule
            FT & 31.10 & 76.59 & 31.28 & 91.24 & 38.98 & 85.62 \\
            \midrule
            ER & {45.89} & {37.42} & {62.44} & {34.77} & {64.96} & {22.29} \\
            \midrule
            A-GEM & 38.23 & 76.59 & 53.36 & 63.52 & 57.34 & 60.34 \\
            \midrule
            EWC & 30.97 & 85.43 & 37.30 & 85.20 & 42.34 & 80.81 \\
            \midrule
            LwF & 24.56 & 69.60 & 45.68 & 68.18 & 46.97 & 54.22 \\
            \midrule
            PB & 21.15 & 59.89 & 37.36 & 78.67 & 37.73 & 66.58 \\
            \midrule
            L2P & 42.91 & 61.60 & 53.15 & 46.46 & 51.97 & 38.99 \\
            \midrule
            \textbf{Double Mixture} & \textbf{51.72} & \textbf{34.33} & \textbf{65.53} & \textbf{24.32} & \textbf{73.12} & \textbf{21.30} \\
            \midrule
            MTL & 60.1 & - & 66.1 & - & 84.5 & - \\
            \bottomrule
        \end{tabular}
    }
    \vspace{.2cm}
    \caption{Main results on three benchmarks. All the results come from our implemented models. The numbers in the table are the average results at the end of a task flow. For each dataset, we mark the best representation in \textbf{bold}.
    Our method consistently outperforms other continual learning baselines on the three datasets.
    }
    \label{tab:res-cl}
\end{table*}


\subsection{Baselines}
We compare our proposed model with the following baselines:
(1) Fine-tuning (FT): We finetuned the Whisper model on several tasks sequentially. 
(2) Multi-task Learning (MTL): We finetuned the Whisper model on several tasks jointly. 
(3) Experience Replay (ER)~\cite{rolnick2019experience}: a rehearsal-based method that leverages a data buffer to store and replay selected experiences from previous tasks.
(4) Averaged Gradient Episodic Memory (A-GEM)~\cite{chaudhry2018efficient}: a rehearsal-based method that extends the concept of ER by enforcing inequality constraints on the loss function for past experiences. 
(5) Elastic Weight Consolidation (EWC)~\cite{kirkpatrick2017overcoming} a regularization-based method that adds regularization on parameters according to their importance to old tasks.
(6) Learning without Forgetting (LwF)~\cite{li2017learning}: a regularization-based method that utilizes a dual-model architecture where a frozen copy of the model acts as a teacher for the current, active model (student). This configuration allows the student model to learn from both the new task and the teacher, enriching its learning trajectory without forsaking prior knowledge.
(7) Piggyback (PB)~\cite{mallya2018piggyback}: an architecture-based method introduces an innovative architectural adjustment that involves the application of binary masks derived from learnable weights onto the frozen parameters of a base model.
(8) Learning to Prompt (L2P)~\cite{wang2022learning} employs a collection of learnable vectors that are dynamically integrated into a pre-trained model.

\subsection{Metrics}
We calculate the average accuracy and average forgetting rate of each method over the entire task sequence.

\noindent \textbf{Average Accuracy.} 
To assess the overall performance across all learned tasks, we calculate incrementally after each newly introduced task:
\begin{equation}
    Avg \; ACC=\frac{1}{T}\sum_{i=1}^{T}R_{T,i},
\end{equation}
where $R_{T,i}$ represents the performance metric on the $i$-th task after training on the $T$-th task.

\noindent \textbf{Forgetting Rate.} 
We quantify the average loss in performance on earlier tasks, denoted as Forgetting Rate:
\begin{equation}
    Avg \; Forgetting=\frac{1}{T}\sum_{i=1}^{T}R_{i,i}-R_{T,i},
\end{equation}
where $Forgetting$ spans the range $(-\infty, +\infty)$, with positive values indicating forgetting on the prior tasks, and negative ones suggesting performance improvement.

\subsection{Implementation Details}
We train the models for 10 epochs per task using AdamW~\cite{loshchilov2017decoupled} with an initial learning rate of 0.0001, and we use a batch size is 16. After each epoch, the learning rate is reduced by 20\% if no validation performance improvement is observed. We clip the gradient L2 norm to 5 to enhance stability. We also freeze Whisper’s encoder and enable automatic mixed-precision to reduce memory consumption and speed up training. We use whisper-base\footnote{\url{https://huggingface.co/openai/whisper-base}} as the backbone and apply greedy decoding at inference. We retain 10\% of the data from each previous task, about 20 to 100 samples. The combination of speech in memory is different from the test set. The gating loss coefficient is set to $0.1$.


\section{Results and Analysis}

\subsection{Main Results}

\begin{table*}[!htbp]
    \centering
    \resizebox{.98\textwidth}{!}
    {
        \begin{tabular}{c|c|c|c|c|c|c|c|c|c|c|c|c|c}
            \toprule
            \multicolumn{13}{c}{\textbf{Speech splicing}} \\
            \midrule
            \textbf{Task} & \textbf{T1} & \textbf{T2} & \textbf{T3} & \textbf{T4} & \textbf{T5} & \textbf{T6} & \textbf{T7} & \textbf{T8} & \textbf{T9} & \textbf{T10} & \textbf{T11} & \textbf{Avg ACC$\uparrow$} & \textbf{Avg Forgetting$\downarrow$} \\
            \midrule
            FT & 86.00 & 44.75 & 32.53 & 19.90 & 22.34 & 16.4 & 20.31 & \cellcolor[HTML]{FABE57}17.78 & \cellcolor[HTML]{FCC95D}15.8  & \cellcolor[HTML]{FED262}14.22 & \cellcolor[HTML]{FFD966}12.93 
            & 27.63 & 68.30  \\
            \midrule
            \textbf{Double Mixture} & 87.00 & 82.75 & 49.73 & 22.90 & 15.74 & 18.13 & 14.55 &
            \cellcolor[HTML]{FEBB03}21.77 & \cellcolor[HTML]{FAAD0E}24.91 & \cellcolor[HTML]{F39222}30.71 & \cellcolor[HTML]{ED7D31}35.04 
            & \textbf{36.57} & 58.89  \\
            \midrule
            w/o Experts & 86.00 & 77.00 & 45.80 & 20.70 & 17.26 & 15.78 & 14.71 &
            \cellcolor[HTML]{FFC000}20.65 & \cellcolor[HTML]{FDB905}22.25 & \cellcolor[HTML]{F7A216}27.27 & \cellcolor[HTML]{F39023}30.96 
            & \underline{34.40} & 66.13   \\
            \midrule
            w/o Memory & 86.00 & 81.75 & 40.20 & 21.12 & 17.93 & 13.73 & 14.37 & 
            \cellcolor[HTML]{FFC000}20.85 & \cellcolor[HTML]{FEBD02}21.5 & \cellcolor[HTML]{F8A414}26.74 & \cellcolor[HTML]{F4951F}29.98
            & 34.02 & 66.47  \\
            \midrule
            \multicolumn{13}{c}{\textbf{Speech overlay}} \\
            \midrule
            \textbf{Task} & \textbf{T1} & \textbf{T2} & \textbf{T3} & \textbf{T4} & \textbf{T5} & \textbf{T6} & \textbf{T7} & \textbf{T8} & \textbf{T9} & \textbf{T10} & \textbf{T11} & \textbf{Avg ACC$\uparrow$} & \textbf{Avg Forgetting$\downarrow$} \\
            \midrule
            FT & 83.00 & 44.75 & 32.53 & 19.90 & 22.34 & 16.4 & 20.31 & 
            \cellcolor[HTML]{FABE57}17.78 & \cellcolor[HTML]{FCC95D}15.8  & \cellcolor[HTML]{FED262}14.22 & \cellcolor[HTML]{FFD966}12.93
            & 27.63 & 71.01 \\
            \midrule
            \textbf{Double Mixture} & 87.00 & 62.50 & 43.60 & 20.46 & 16.00 & 15.57 & 14.40 & \cellcolor[HTML]{F9B752}19.04 & \cellcolor[HTML]{F7B04F}20.13 & \cellcolor[HTML]{F1923D}25.48 & \cellcolor[HTML]{ED7D31}29.05
            & \textbf{31.75} & 61.09 \\
            \midrule
            w/o Experts & 85.00 & 47.25 & 27.60 & 21.01 & 16.76 & 15.86 & 14.45 & 
            \cellcolor[HTML]{FABC56}18.04 & \cellcolor[HTML]{FBC159}17.16 & \cellcolor[HTML]{F8B24F}19.93 & \cellcolor[HTML]{F6A84A}21.55
            & 27.39 & 68.71  \\
            \midrule
            w/o Memory & 85.00 & 47.00 & 41.00 & 19.91 & 16.76 & 15.97 & 13.44 & \cellcolor[HTML]{FCC75C}16.19 & \cellcolor[HTML]{FCC95D}15.9 & \cellcolor[HTML]{F9B954}18.69 & \cellcolor[HTML]{F7AF4E}20.38
            & \underline{28.20} & 68.30  \\
            \bottomrule
        \end{tabular}
    }
    \caption{Performance on incremental tasks in Speech splicing and Speech overlay datasets. T1-T7 are tasks in which single semantic and acoustic events are randomly ordered, and T8-T11 are tasks in which two types of events are combined in speech. The latter four tasks are used to evaluate the generalization ability of the model, and their instances are not included in the training set.
    Given that other continual learning baseline methods can only predict single events and cannot identify combined events, they are not recorded in the table.
    }
    \label{tab:res-cg}
\end{table*}


\subsubsection{Comparison of Continual Learning Methods}
Table~\ref{tab:res-cl} reports the average ACC and average Forgetting metrics at the end of the curriculum for different methods, and we observe that: 
(1) our method achieves the best performance, with the proposed method achieving 51.72\% accuracy on the Speech ACE05, compared to 45.89\% for the replay-based ER method and only 30.97\% for the regularization-based EWC method. This result emphasizes the advantages of the proposed method in balancing new knowledge learning with old knowledge retention.
(2) The advantages shown by the replay-based methods (ER and A-GEM) may stem from their ability to directly utilize historical data for retraining, where ER achieves 45.89\% on the Speech ACE05 dataset, which is significantly higher than that of the EWC (30.97\%) and LwF (24.56\%). This reveals the effectiveness of the replay mechanism in promoting model memory retention.
(3) Regularization-based methods perform poorly in the experiments, e.g., the accuracy of EWC is only 30.97\%. This may be because while regularization constraints can help the model retain old knowledge, they may also limit the model's learning of new tasks, especially when faced with new tasks that differ significantly from previous tasks.
(4) Dynamic architecture methods (e.g., PB and L2P) provide the ability to dynamically adjust the model structure but still fall short in terms of forgetting rate, e.g., the forgetting rate of PB on Speech ACE05 is 59.89\%, which is higher than that of the proposed method, which is 34.33\%. This implies that although the dynamic architecture can introduce new network structures for new tasks, there are still challenges in long-term memory retention.

\subsubsection{Comparison of Backbone Model}
We compare the performance of some baseline methods using Whisper and WavLM as backbone models. Note that WavLM is an Encoder-only architecture, and we concatenate an LSTM as a text decoder after its speech Encoder~\cite{yang2023investigating}. As shown in Table~\ref{tab:res-backbone-model}, we observe that Whisper performs better than WavLM in all three methods. Moreover, WavLM cannot identify acoustic events in the ESC-50 dataset. Therefore, we use Whisper in subsequent experiments.
\begin{table}[!htbp]
    \centering
    \resizebox{.37\textwidth}{!}{
    \begin{tabular}{c|c|c}
        \toprule
        \textbf{Backbone} & \textbf{Whisper-base} & \textbf{WavLM-base} \\ 
        \midrule
        FT & $31.10$ & $21.03$ \\ 
        ER & $45.89$ & $28.3$ \\ 
        EWC & $30.97$ & $20.72$ \\ 
        Double Mixture & $51.72$ & -- \\ 
        \hline
        MTL & $60.1$ & $39.6$ \\ 
        \bottomrule
    \end{tabular}
    }
    \caption{Average accuracy (\%) on Speech ACE05. Note that WavLM, being an encoder-only architecture, cannot implement our method as it requires a decoder.
    }
    \label{tab:res-backbone-model}
    \vspace{-.5cm}
\end{table}

\subsubsection{Exploration of Event Disentanglement}
We further evaluate the event disentanglement ability of the proposed method, and the experimental results are summarized in Table \ref{tab:res-cg}. 
We find that 
(1) On the SS dataset, the proposed method performs better in terms of average accuracy compared to the FT method and the ablation variants and similarly shows the same trend on the SO dataset. 
(2) The ablation experiments show that removing either the expert group or the memory mechanism leads to performance degradation, which confirms the importance of these two components in our method.
(3) From the results, it can be observed that the average accuracy of Speech splicing is generally higher than that of Speech overlaying, which may be because the superimposed speech samples have more interference in the acoustic signal, which increases the difficulty of the model in recognizing and distinguishing different events. In addition, the interweaving of acoustic elements in the overlay samples may lead to more information loss, whereas the spliced examples maintain more information about the context of the event, providing a clearer basis for the model to make judgments.

\subsection{Ablation Studies}
\subsubsection{Importance of Each Component}
\begin{figure}[!htbp]
  \centering
    \includegraphics[width=0.4\textwidth]{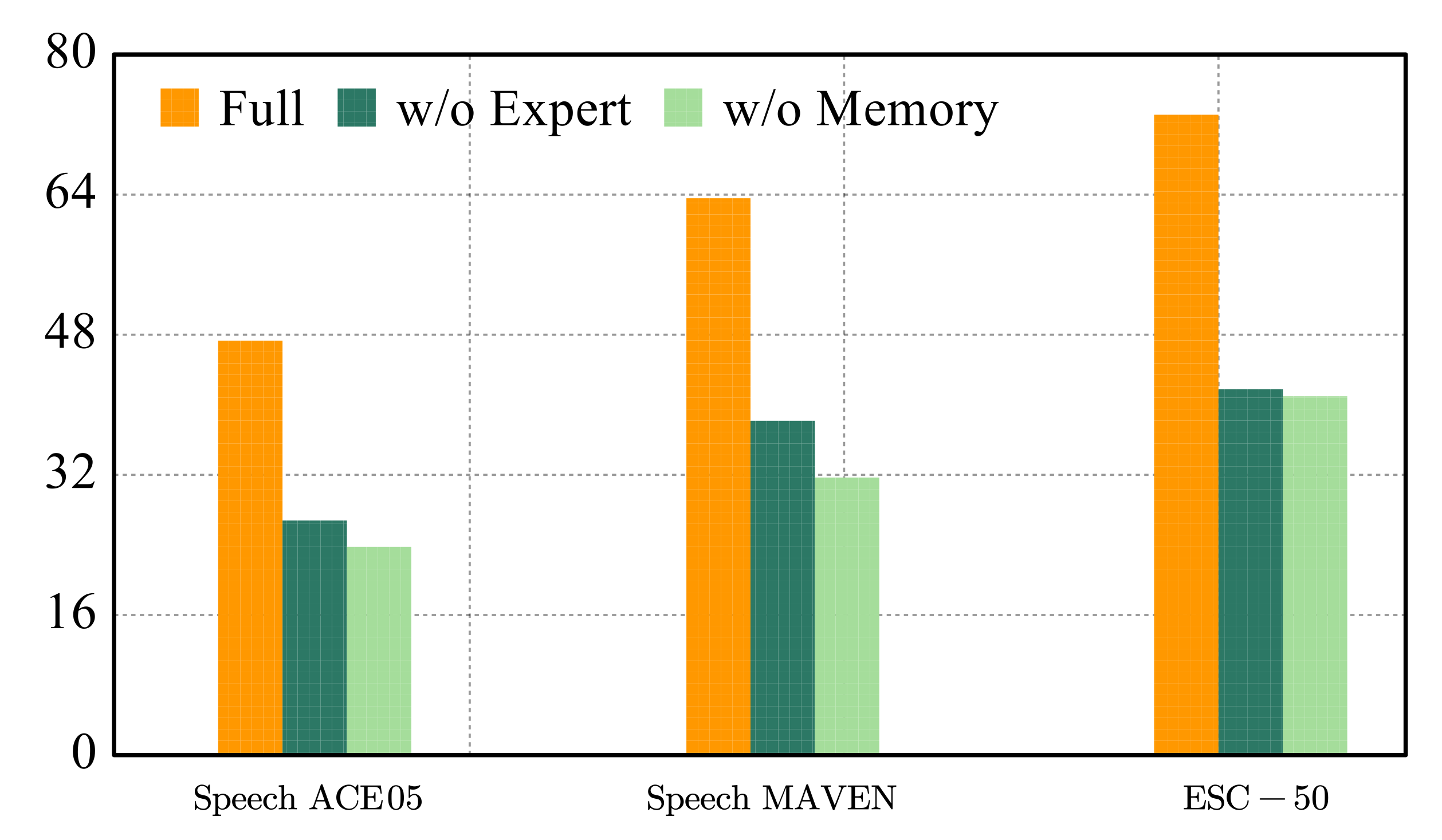}
    \caption{Ablation study on three datasets, the horizontal axis represents different data sets, and the vertical axis represents the average accuracy of the entire task sequence.
    }
    \Description{ToDo}
    \label{fig:res-ablation}
\end{figure}

As can be seen in Figure~\ref{fig:res-ablation}, the accuracy of model recognition events decreases when speech experts are removed; this leads to a loss of about ~21\% in Speech ACE05, ~25\% in Speech MAVEN, and ~32\% in ESC-50. Similar findings are seen throughout the examined datasets, indicating that the model's performance is significantly harmed by memory reduction. 



\subsubsection{Impact of Task Order}
We investigate the impact of task ordering with our method on the ESC-50 datasets(See Table \ref{tab:data-task-order-esc} and \ref{tab:res-task-order-esc}) and Speech MAVEN (See Table \ref{tab:data-task-order-maven} and \ref{tab:res-task-order-maven}).
Results show that different task orders resulted in differences in overall accuracy. For example, 
sequence $\#2$ has an average accuracy of 73.06\%, but sequence $\#4$ yields an accuracy of 69\% (in ESC-50). 
This demonstrates that even small adjustments to task order can affect model generalization and the ability to apply learned knowledge.
However, our method still outperforms other baseline methods at the worst performance level.
Similar observations can be found in the Speech MAVEN dataset.

\begin{table}[!htb]
    \centering
    \resizebox{0.45\textwidth}{!}{
        \begin{tabular}{c|c}
            \toprule
            \textbf{Order} & \textbf{Task Sequence} \\ 
            \midrule
            \#1 & $ animal \rightarrow domestic \rightarrow human \rightarrow nature \rightarrow urban $  \\ 
            \#2 & $ domestic \rightarrow nature \rightarrow human \rightarrow urban \rightarrow animal $  \\ 
            \#3 & $ domestic \rightarrow urban \rightarrow nature \rightarrow animal \rightarrow human $  \\ 
            \#4 & $ animal \rightarrow urban \rightarrow nature \rightarrow human \rightarrow domestic $  \\ 
            \#5 & $ domestic \rightarrow urban \rightarrow animal \rightarrow human \rightarrow nature $  \\ 
            \bottomrule
        \end{tabular}
    }
    \caption{Random different task sequences on ESC-50.}
    \label{tab:data-task-order-esc}
\end{table}
\vspace{-.55cm}

\begin{table}[!htb]
    \centering
    \resizebox{0.4\textwidth}{!}{
        \begin{tabular}{c|c|c|c|c|c|c}
            \toprule
            \textbf{Order} & \textbf{T1} & \textbf{T2} & \textbf{T3} & \textbf{T4} & \textbf{T5} & \textbf{Avg ACC} \\
            \midrule
            \#1 & 81.00 & 76.85 & 67.87 & 69.38 & 66.52 & 72.32  \\
            \#2 & 85.00 & 70.00 & 72.10 & 69.40 & 68.82 & 73.06  \\
            \#3 & 85.00 & 71.25 & 62.00 & 69.47 & 67.50 & 71.04  \\
            \#4 & 81.00 & 72.60 & 68.27 & 64.12 & 59.02 & 69.00  \\
            \#5 & 85.00 & 71.25 & 62.90 & 64.93 & 64.94 & 69.80  \\
            \bottomrule
        \end{tabular}
    }
    \caption{Average accuracy (\%) on incremental tasks in ESC-50.
    }
    \label{tab:res-task-order-esc}
\end{table}
\vspace{-.55cm}


\begin{table}[H]
    \centering
    \resizebox{.45\textwidth}{!}{
        \begin{tabular}{c|c}
            \toprule
            \textbf{Order} & \textbf{Task Sequence} \\ 
            \midrule
            \#1 & $ process \rightarrow action \rightarrow scenario \rightarrow move \rightarrow life \rightarrow talk $  \\ 
            \#2 & $ action \rightarrow process \rightarrow move \rightarrow life \rightarrow scenario \rightarrow talk $  \\ 
            \#3 & $ scenario \rightarrow process \rightarrow action \rightarrow move \rightarrow life \rightarrow talk $  \\ 
            \#4 & $ move \rightarrow talk \rightarrow action \rightarrow process \rightarrow life \rightarrow scenario $ \\  
            \bottomrule
        \end{tabular}
    }
    \caption{Random different task sequences on Speech MAVEN.}
    \label{tab:data-task-order-maven}
\end{table}
\vspace{-.55cm}

\begin{table}[H]
    \centering
    \resizebox{.4\textwidth}{!}{
    \begin{tabular}{c|c|c|c|c|c|c|c}
        \toprule
        \textbf{Order} & \textbf{T1} & \textbf{T2} & \textbf{T3} & \textbf{T4} & \textbf{T5} & \textbf{T6} & \textbf{Avg ACC} \\ 
        \midrule
        \#1 & 86.00 & 74.40 & 64.00 & 57.10 & 52.46 & 51.67 & 64.27 \\
        \#2 & 76.00 & 72.35 & 61.90 & 56.50 & 54.52 & 53.23 & 62.42 \\
        \#3 & 82.00 & 69.20 & 65.10 & 55.78 & 53.18 & 52.87 & 63.02 \\
        \#4 & 79.00 & 70.35 & 62.33 & 61.65 & 55.34 & 50.97 & 63.27 \\
        \bottomrule
    \end{tabular}
    }
    \caption{
    Average accuracy (\%) on incremental tasks in Speech MAVEN.
    }
    \label{tab:res-task-order-maven}
\end{table}
\vspace{-.3cm}


\section{Related Work}

\subsection{Event Detection}

Event Detection has been studied mostly focused on textual data, involving feature-based models~\cite{yang2016joint} and deep learning models~\cite{yang2019exploring,wang2023continual,WuSKQHL23}. Recent work investigates the disparities between extracting events from text and speech, introducing an end-to-end approach for speech event extraction~\cite{kang2024event}. Conventional acoustic event detection methods~\cite{bilen2020framework,martin2023training,xu2023semi} are typically characterized by a three-fold approach: (1) leveraging convolutional neural networks to extract acoustic features either from spectrogram representations or directly from audio waveforms; (2) employing recurrent neural networks or Transformers alongside a mean or max pooling mechanism to delineate temporal relationships among frame-level features; and (3) calculating the Connectionist Temporal Classification objective function.

\subsection{Continual Learning}
Existing continual learning algorithms~\cite{abs-2402-01364,abs-2401-16386,abs-2302-00487} can be categorized into three main groups: 
(1) \textit{Rehearsal-based methods}~\cite{rolnick2019experience, lopez2017gradient, chaudhry2018efficient} involve either storing a small subset of training samples from previous tasks in memory or employing a data generator to produce pseudo samples of past tasks. When learning a new task, both retained and generated samples, along with new task data, are utilized for training. 
(2) \textit{Regularization-based methods}~\cite{kirkpatrick2017overcoming, li2017learning} apply penalties or regularization to discourage changes to important parameters acquired from previous tasks when learning a new task. 
(3) \textit{Architecture-based methods}~\cite{mallya2018piggyback, serra2018overcoming, wang2022learning} allocate distinct parameters to different tasks to prevent interference with previously learned parameters, necessitating task identification during both training and testing. 
%
While these techniques mainly originate from computer vision~\cite{abs-2305-08698} and are widely used in natural language processing~\cite{WuCLLQH22,YuJN21,LiuCH22}, there is a noticeable research gap on continual speech learning, with that limited studies~\cite{della2023cl, vander2022continual,YangDCWW22,ShamsiLBMPTMPGG23} have explored continual learning for monolingual and multilingual ASR tasks. 

\section{Conclusion}
This paper introduces a new task designed to continually extract both semantic and acoustic events from speech, focusing on overcoming catastrophic forgetting and improving event disentanglement. We propose a novel approach called "Double Mixture," employing a combination of Mixture of Experts and Mixture of Memory mechanisms. This strategy effectively mitigates catastrophic forgetting and enhances the model’s ability to generalize across different types of events, outperforming existing continual learning benchmarks in terms of accuracy and forgetting metrics. Future work will explore applying this approach to broader speech processing and understanding tasks.

\bibliographystyle{ACM-Reference-Format}
\bibliography{mybib}

\end{document}